\documentclass[preprint,12pt]{elsarticle}

\usepackage{amssymb}
\usepackage{xurl}
\usepackage[utf8]{inputenc}
\usepackage[T1]{fontenc}

\usepackage{booktabs} 
\usepackage{multirow} 
\usepackage{xcolor}
\usepackage{fancyvrb}
\usepackage{subcaption}
\usepackage{float}

\journal{AI\&Law}

\begin{document}

\begin{frontmatter}

\title{ChatGPT Unveils Its Limits: Principles of Law Deliver Checkmate}

\author[inst1,inst2,inst3]{Marianna Molinari}
\author[inst2]{Ilaria Angela Amantea}
\author[inst1,inst2]{Marinella Quaranta}
\author[inst5]{Guido Governatori}

\affiliation[inst1]{organization={Univeristy of Bologna - LaST-JD, Legal Studies Dept., Italy},
            }

\affiliation[inst2]{organization={University of Turin - Computer Science Dept., Italy},
            }
            
\affiliation[inst3]{organization={Vrije Universiteit Brussel - PREC Dept., Belgium},
            }

\affiliation[inst5]{organization={School of Engineering and Technology, Central Queensland University, Australia},
            }

\begin{abstract}
This study examines the performance of ChatGPT with an experiment in the legal domain. We compare the outcome with it a baseline using regular expressions (Regex), rather than focusing solely on the assessment against human performance. The study reveals that even if ChatGPT has access to the necessary knowledge and competencies, it is unable to assemble them, reason through, in a way that leads to an exhaustive result. This unveils a major limitation of ChatGPT. Intelligence encompasses the ability to break down complex issues and address them according to multiple required competencies, providing a unified and comprehensive solution. In the legal domain, one of the most crucial tasks is reading legal decisions and extracting key passages condensed from principles of law (PoLs), which are then incorporated into subsequent rulings by judges or defense documents by lawyers. In performing this task, artificial intelligence lacks an all-encompassing understanding and reasoning, which makes it inherently limited. Genuine intelligence, remains a uniquely human trait, at least in this particular field.
\end{abstract}


\end{frontmatter}

\section{Introduction}

Legal professionals, including judges and lawyers, often face challenges stemming from limited access to IT support and insufficient technical proficiency. These issues are especially prominent during the ongoing generational transition within the legal sector, which demands increased adaptability to technological advancements.

Artificial intelligence tools, such as chatbots, are designed to be user-friendly and accessible, even for individuals with limited technical expertise. These systems possess contextual understanding, enabling them to process both legal inquiries and factual questions effectively. By analyzing conversational flow and prior interactions, AI-driven chatbots can generate more relevant and personalized responses, thereby improving user engagement and operational efficiency. Recent academic research suggests that these capabilities are increasingly applicable to the legal domain.

ChatGPT, the AI model employed in this study, demonstrates a strong ability to generate human-like responses with precise grammar, interpretive comprehension, and adaptability, while maintaining a conversational tone. Its intuitive interface, free accessibility, and lack of installation requirements make it particularly well-suited for small courts and law firms, which often operate with limited technology budgets.

While this theoretical framework has been established, it is essential to empirically validate and assess ChatGPT's ability to perform legal tasks. From this perspective, the evaluation should extend beyond a singular, self-referential assessment and instead incorporate a comparative analysis. This approach enables a more comprehensive understanding of the model’s strengths and limitations within the legal domain.

When evaluating LLMs, it is crucial to compare their work with other methods to determine whether they perform better or worse. Essentially, it is necessary to attempt falsification \cite{popper1963science}. Thus, a reference baseline is required. Additionally, assessing their capabilities relative to human results is crucial for understanding their practical utility.

This study evaluates ChatGPT's effectiveness in extracting principles of law (PoLs), using regular expressions (Regex) as a baseline to highlight its limitations. Notably, this baseline relies on straightforward and accessible techniques, ensuring that the falsification process remains reproducible and verifiable.

Section 2 provides background information and outlines the study’s methodology. Section 3 details the experimental setup, including the dataset, the guidelines, and the limits encountered by expert annotators, along with the results. Section 4 examines the prompting phase, focusing on ChatGPT's prompting strategy and the process of identifying PoLs. Section 5 presents a comparative analysis of the findings obtained by expert annotators and ChatGPT. Section 6 discusses the development of a Regex using ChatGPT and its implementation within a text editor for PoLs extraction. Section 7 compares the results obtained by expert annotators with those extracted using Regex. Section 8 offers a final comparison of the extractions performed by expert annotators, ChatGPT, and Regex. Based on the previous comparisons, section 9 questions whether AI is truly intelligent. Lastly, section 10 ends with conclusions and directions for future works.

\section{Background and Methodology} 

PoLs provide a fundamental framework for interpreting and applying legal rules to specific cases. In doing so, PoLs contribute to the universalization of individual decisions \cite{fidelbo2018verso}.

Typically, PoLs emerge from the specific facts of a given case and gradually evolve into more general statements \cite{guastini2004interpretazione}. If confined to individual cases, PoLs would lack the necessary generality to ensure legal certainty in future applications. Instead, they serve as interpretive guidelines for judges, fostering predictability in judicial decisions, consistency within the legal system, and a desirable deflationary effect. Thus, these principles are defined to apply across an indefinite number of cases and can also play a crucial role in resolving disputes beyond the courtroom.

The development of a method for the automatic extraction of PoLs is therefore essential. This approach prioritizes practical outcomes over purely theoretical or philosophical legal considerations \cite{amantea2021semi, amantea2019modelling, prakken2024evaluating}. By emphasizing accuracy, it seeks to enhance the efficiency of PoLs identification and application, thereby facilitating their role in case resolution \cite{guastini1998principi, colombino2023organizing, alfred2014algorithms}.

In this paper, we unveil the limits of ChatGPT in the execution of the task.

ChatGPT has shown constraints in the automatic, straightforward extraction of PoLs from judgments. Its current performance remains insufficient in both the quantity and accuracy of extracted PoLs, underscoring its shortcomings in basic legal research.

This limitation is demonstrated in this paper through an empirical-comparative approach.

The study outlines the methods used to extract PoLs through ChatGPT and presents an analysis of the obtained results. These are initially compared with those produced by expert annotators to assess the model’s effectiveness.

Through previous research, we have already explored preliminary attempts at automating PoLs extraction, using different techniques \cite{zeugmann2011precision, rijsbergen1979information, pedregosa2011scikit}. Our investigation began with basic experiments, leveraging regular expressions (Regex) directly generated by domain experts to extract PoLs from Italian judicial decisions. Early findings revealed that PoLs do not adhere to a consistent syntactic structure, making their automated extraction challenging. However, citations of Italian Supreme Court rulings demonstrated to be patterns more clearly defined and less ambiguous compared to other sources \cite{MariICAIL2023, MariITAIS2023}.

Building on these observations, a subsequent study explored the use of ChatGPT for the automatic extraction of PoLs \cite{MariJURIX2024}. While this approach enabled the identification of some PoLs, the results proved suboptimal in both quantity and quality. ChatGPT demonstrated the ability to provide general guidance within a judgment but often generated incorrect extractions. These errors were particularly problematic as they frequently involved subtle misinterpretations - such as the use of misleading synonyms - that preserved the surface appearance of the content while deviating from its original meaning. This issue made errors difficult to detect, even for legal experts.

To further evaluate these findings, a retrospective analysis is here conducted using a baseline relying on Regex. In this case, the regular expressions were generated directly by ChatGPT and applied to the dataset using a simple text editor. The results of this approach were then subjected to a second comparative evaluation, again measuring performance against expert annotations.

Finally, the study presents a comprehensive comparative assessment of all three extraction methods - expert annotators, ChatGPT, and Regex - providing a systematic evaluation of their respective strengths and limitations.

\section{Expert-Led Annotation Process}

\subsection{Dataset}

PoLs were extracted from a corpus of Italian judgments concerning LGBTQIA+ rights, specifically rulings issued by first and second-instance courts (Tribunals and Courts of Appeal). These judgments were randomly selected in .docx format from De Jure database\footnote{De Jure is an advanced online legal information system with semantic search capabilities. Access to De Jure is provided to institutional users via the Italian University Library System (SBA) (https://dejure.it). All rights reserved by Giuffrè Francis Lefebvre: the publisher provided consent for the use of the data in the present research only. Therefore, the text of the judgments cannot be shared.}. The selection process was conducted by filtering judgments using the keyword \textit{“same-sex”}.

In the Italian legal system, cases involving same-sex relationships and LGBTQIA+ rights often encounter a lack of clear legislative guidance. In such contexts, PoLs serve a crucial function in bridging this gap. Notably, the first law addressing civil unions for same-sex couples and \textit{de facto} cohabitation was only enacted in 2016 (Law No. 76, May 20, 2016), yet it remains incomplete on certain issues, such as stepchild adoption. These judgments were therefore selected as the foundation of the dataset, as they provide a valuable source of PoLs for analysis and extraction.

A total of 60 judgments, ranging from 3 to 38 pages in length, were randomly retrieved. The dataset comprises 7 judgments from Courts of Appeal, 50 from Tribunals, and 3 from Juvenile Courts, all originating from different Italian judicial districts. These judgments were categorized into two groups: originals and copies. The original judgments were assigned to a team of three legal domain experts for annotation, while the copies were processed separately. Each expert was assigned 20 judgments from the total collection of 60, following the annotation guidelines outlined below.

\subsection{Guidelines}

The \textit{corpus}, in its original .docx format, was provided to a group of annotators with expertise in both law and research methodology. The manual annotation should be performed by legal professionals, as it demands specialized legal knowledge, even when working with quoted material. For instance, a mention of a Supreme Court ruling holds far greater importance than a technical report or an expert's opinion. Additionally, these references might be presented without quotation marks or an explicit citation, as explained later \textit{infra}. Therefore, legal expertise was crucial to ensure that the annotations were both accurate and contextually relevant.

Each annotator was initially assigned a set of 20 judgments for annotation. The annotated judgments were then reviewed and validated by the other two annotators. The annotation process adhered to predefined guidelines, collaboratively established and shared among the annotators. They opted for a single reading of each judgment, manually highlighting the PoLs within the .docx file using different colors to distinguish them

Direct PoLs - those enclosed in quotation marks (e.g., English quotation marks or high single marks (`…'), Italian quotation marks or high double marks (“…”), French marks or low double marks («…»), etc.) - were highlighted in yellow.
In contrast, indirect PoLs, which paraphrase the referenced passage without the use of quotation marks, were highlighted in blue.

Both direct and indirect PoLs are classified as explicit PoLs, as they are explicitly connected to a specific prior citation from a Judicial Office. In contrast, PoLs with no clear link to a specific prior citation were highlighted in gray and categorized as implicit PoLs.

\subsection{Limits}

Upon completing the manual annotation, the annotators were asked to summarize the challenges encountered during the process. They reported the following:

\begin{itemize}
    \item Judgments are typically drafted in varying styles, which influence both the wording and the presentation of different types of PoLs and their corresponding citations.
    \item Explicit direct PoLs may appear within various types of quotation marks (e.g., English quotation marks or high single marks (`…'), Italian quotation marks or high double marks (“…”), French marks or low double marks («…»), etc.). They may or may not be preceded by a colon and are often introduced by phrases such as `it has been established,' `it has been affirmed,' `it has been thus decided,' or similar expressions.
    
    \item Explicit indirect PoLs and implicit PoLs may be paraphrased in various ways, depending on the writing style. They may or may not be introduced by the aforementioned expressions.
    
    \item Citations may appear before or after explicit PoLs, with or without parentheses. They can also be introduced by keywords such as `cfr.' (`compare') or `v.' (`see') and may be presented in full or abbreviated forms. For instance, when citing explicit PoLs from the Italian Supreme Court of Cassation, the judgment identifier – comprising a code and year (e.g., Cass.~n.~26972/2008) – is commonly used. Alternatively, citations may include additional details, such as the issuing judicial section and the date, in formats such as: (i) Civ.~Cass., UU.~SS., n.~26972/2008; (ii) Court of Cassation, United Sections, n.~26972 of November 11, 2008.
    
    \item Quoted passages may include not only PoLs, but also references to statutory provisions or narrative excerpts.
\end{itemize}

\subsection{Results}

Following the expert-led annotation process, Table \ref{ann} was created to document the total number of PoLs identified from the pool of 60 annotated judgements, and their distribution by typology.  A total of 682 PoLs were identified: 87 implicit, 293 explicit-direct, and 302 explicit-indirect. 

\begin{table}[h!]
    \centering
    \begin{tabular}{ccccc}
\textbf{Judgments}  & \textbf{Total PoLs} & \textbf{Implicit} & \textbf{Direct} & \textbf{Indirect} \\
\midrule
60 &  682
& 87&293	&302
    \end{tabular}
    \caption{PoLs Annotation by Legal Domain Experts.}
    \label{ann}
\end{table}

These results served as the initial comparative baseline for assessing the performance of automatic extraction methods using ChatGPT and Regex.

\section{Exploration with ChatGPT}

The exploration with ChatGPT was conducted in two phases: (i) discovering prompts and (ii) identifying PoLs\footnote{For this experiment, the publicly available version of ChatGPT, accessible via browser, was used in September 2024.}.

\subsection{Discovering prompts}

The prompt discovery phase aimed to find the best input for ChatGPT to extract all PoLs from a specific judgment. This initial exploration was carried out on a small dataset of five judgments.

The initial approach involved “Zero-shot learning”, where ChatGPT received only a textual explanation of the task without examples. Relying on prior knowledge, it generated responses based on the given prompt (Prompts 1 and 2). However, this method had limited success. To improve performance, the study adopted “Few-shot learning”, which included relevant examples from the annotated dataset in the prompt (Prompts 3 and 4) \cite{yu2023exploring}.

Here is a selection of the most relevant prompts, each tested in different versions\footnote{Although ChatGPT performs better in English, the prompts were provided in Italian to align with the original sources – Italian court rulings – and preserve the specific nuances of national legal terminology. The passage presents the English translation of the different prompts tested. The original Italian text is provided in the Appendix.}.

\begin{enumerate}
    \item `Extract the paragraphs in which there is a passage in quotation marks'.
    \item `Extract the paragraphs in which there is the word COURT and similar followed by a passage in quotation marks and a citation in parenthesis'.
    \item `Extract the paragraphs in which there is the word COURT and similar followed by a passage in quotation marks and a citation in parenthesis (e.g. Court of Cassation, sentence no.~15138/2015; Cass.~15138/2015; Cass.~Civil U.S. no.~12193/2019; Cass., sect.~I, 22/06/2016 no.~12962; sent.~22.06.2016 n.~12962; and similar)'.
    \item `Extract the paragraphs in which there is the word COURT, TRIBUNAL, JURISPRUDENCE, COLLEGE, CONSESSION, CASSATION and similar followed by a passage in quotation marks.
Example: The Court of Cassation itself, ruling in United Sections, recently stated that these are cases “which profoundly question individual and collective conscience, posing delicate and complex questions, susceptible to differentiated solutions”.

Also paragraphs where there is a passage in quotation marks followed by a number in parenthesis.
Example: These are cases ``that deeply question individual and collective conscience, posing delicate and complex questions, susceptible to differentiated solutions” (see Cass.~Civil U.S.~n.~12193/19).

Also paragraphs until a new one where the expressions `jurisprudence has held that...', `as the court has ruled...', `established jurisprudence...', `the principle established by the court...', followed or preceded by a number.
Examples:

- The Constitutional Court, in sentence no.~120/2001, has clearly stated that the name, understood as the first and immediate distinctive sign, constitutes one of the inviolable rights of the person protected by the Charter pursuant to art.~2 of the Constitution, which is recognized as an open clause, with the consequent possibility of deducing, from the combined reading of the art.~6 c.c., paragraph 3, and articles 2 and 22 of the Constitution, the nature of an irrepressible subjective right of the person.

- The recognition of the primary right to sexual identity, underlying the required rectification of the attribution of sex, makes the rectification of the first name consequential, which does not necessarily have to be converted into the gender resulting from the rectification, as the judge must take into account the new first name, indicated by the person, even if completely different from the previous name, where this indication is legitimate and compliant with the new status (Cass.~Civ.~3877/2020).

- The expenses of the ctu, in the amount paid by separate decree and the halving provided for by the art.~130 tusg, must be paid by the Treasury (Cost.~Court 217/2019).

- The Court, after stating that the legislator [...], adds: ``the legislator [...]”.
Also all phrase that end with a number in brackets before the full stop.
Examples: ‘…(Cass.~Civ.~3877/2020).’ Or: ‘…(Cost.~Court 217/2019).’

EXACTLY COPY THE PASSAGES FROM THE SINGLE UPLOADED FILE. EXACTLY EXPORT THEM AS THEY ARE FROM THE SINGLE UPLOADED FILE. DO NOT INVENT. DO NOT SUMMARIZE. DO NOT ASSEMBLE.

Follow the instructions in detail.' 
\end{enumerate}

The final prompt was selected as the most suitable for the analysis. Earlier versions were too simplistic and lacked detail, leading to incomplete extractions of PoLs. They often misclassified implicit and explicit-indirect principles as explicit-direct ones and occasionally introduced quotation marks or references absent from the original text. To address these issues, a more precise and refined prompt was developed.

From a technical standpoint, a detailed prompt may appear more effective than a broad one. However, in this study, the choice was guided by the needs of the target users. Judges and lawyers prioritize efficiency, aiming to identify all PoLs in a judgment as quickly as possible. This approach enables them to fully understand the precedent and determine which principles to apply in their rulings and legal acts.

\subsection{Identifying PoLs}

The identification process aimed to use ChatGPT to detect PoLs previously marked by the annotators. Two experts managed this task. One executed queries in ChatGPT, inputting the selected prompt along with each judgment in .docx format, reviewing the results, and reading them aloud.

The second expert reviewed the original formatted versions used by the annotators and supervised the comparison process. They verified whether the PoLs identified by ChatGPT and read aloud by the first expert matched those manually annotated. Additionally, they recorded the findings in an .xlsx file, following the criteria detailed in the next section.  

This procedure was repeated 60 times, once for each annotated judgment.

\section{Analysis and Results: Annotators vs ChatGPT}

\subsection{Analysis}

\begin{figure*}[ht!]
\centering
\includegraphics[width=1\textwidth]{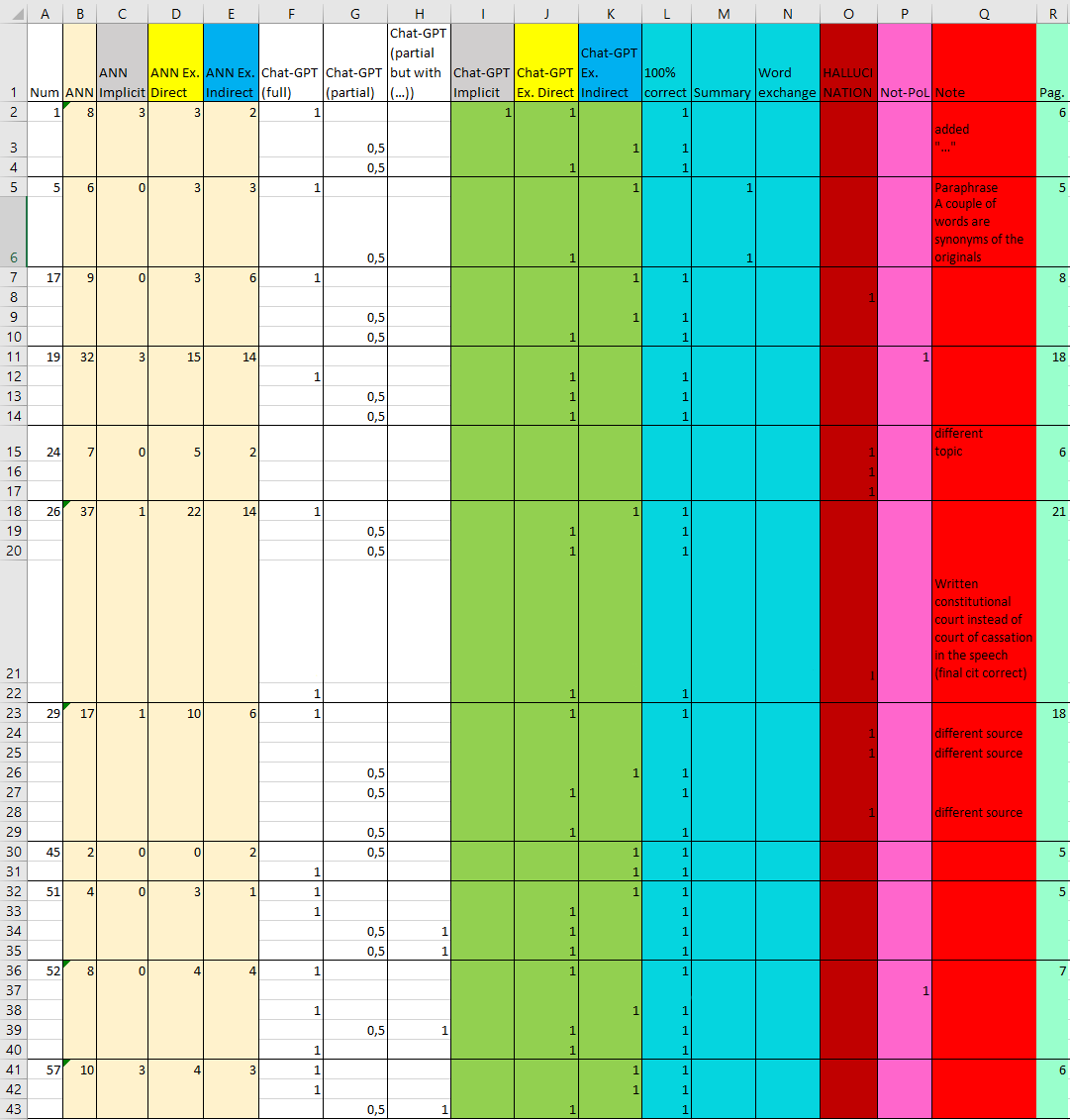}
\caption{Excerpt of the annotation table tracking ChatGPT's PoLs extraction.}\label{tabAnn}
\end{figure*}

Figure \ref{tabAnn} shows an excerpt from the table used to track the annotation process. It illustrates the comparison between the PoLs identified by the annotators and those detected by ChatGPT.

The beige columns (B-E) represent the PoLs identified by the annotators, categorized as follows: the total number of detected PoLs (“ANN”), implicit PoLs (“ANN Implicit”), explicit direct PoLs (“ANN Ex.\ Direct”), and explicit indirect PoLs (“ANN Ex.\ Indirect”). In contrast, the white (F-H) and green (I-K) columns display the PoLs detected by ChatGPT.

Unlike the annotators' columns, ChatGPT's results are further divided into additional columns to enable a more detailed and comprehensive analysis.

Specifically, the white columns indicate whether the PoLs were fully captured (“Chat-GPT (full)”) or only partially identified. Among the partial cases, two distinct patterns emerged: some PoLs were cut off at a full stop (“.”), while others were truncated mid-sentence with an ellipsis (“...”).

These cases were distinguished as they may indicate different scenarios. In the first case, the PoL appears to be fully identified but only partially reported (“Chat-GPT (partial)”). In the second one, the presence of “(...)” suggests that the PoL may extend beyond the extracted portion (“Chat-GPT (partial with (...))”), possibly due to character limitations in ChatGPT's responses.

The green columns classify the PoLs identified by ChatGPT, following the same structure as the beige columns for human annotations. Specifically, they represent implicit PoLs (“Chat-GPT Implicit”), explicit direct PoLs (“Chat-GPT Ex.\ Direct”), and explicit indirect PoLs (“Chat-GPT Ex.\ Indirect”).

The blue columns (L-N) indicate how ChatGPT reported the PoLs. Specifically, they show whether each PoL was reproduced exactly as in the original text (“100\% correct”), summarized or rephrased with synonyms while preserving its meaning (“Summary”), or altered with one or more word substitutions that changed the substance of the statement (“Word exchange”).

The dark red (O) and pink (P) columns indicate errors, specifically hallucinations (“HALLUCINATION”) and false PoLs generated by ChatGPT (“Not-PoL”). Additionally, there is a column for notes on the extraction process (“Note”), and a column indicating the length of the analyzed judgments, measured in pages (“Pag.”). 

The analysis shows that the query performs best when there are 2 to 4 PoLs, with the highest accuracy occurring at exactly 3 PoLs (e.g., Judgments No.~45 and 51). However, as the number of PoLs increases, accuracy declines (e.g., Judgment No.~26). Regardless of the total PoLs or text length, the query consistently detects an average of 3 to 5 PoLs.  
The analyzed judgments ranged from 3 to 38 pages, but sentence length does not appear to affect PoL identification accuracy. For example, while the query correctly identified all PoLs in the 5-page Judgments No.~45 and 51, it failed in the similarly short Judgment No.~5. Likewise, in the 6-page Judgment No. 57, it detected only 3 out of 10 PoLs.

This observation highlights two key issues. First, readers cannot determine the exact number of PoLs in a text and can only assume that longer judgments are more likely to contain more PoLs. Second, even when the exact number of PoLs is known, ChatGPT struggles to identify more than a limited subset.  For example, in a test with a 38-page judgment containing 55 PoLs, ChatGPT initially identified only 3 (1 correct and 2 incorrect). When prompted to provide all existing principles, it added just one more PoL (also incorrect). Finally, even when explicitly instructed to return all 55 PoLs, it identified only 10.

Additionally, explicit direct PoLs are the easiest to identify due to their consistent and recognizable structure. In contrast, implicit PoLs, which lack a clear composition, are significantly more challenging to detect.

Furthermore, in most cases, the extracted text matches the source exactly, as shown by the predominance of the “100\% correct” column over the other blue columns. However, this is not always the case. Occasionally, the text is slightly summarized, reordered (e.g., in lists), or modified with synonyms. These changes typically occur with explicit direct PoLs, where minor adjustments - such as synonym substitutions - do not alter the meaning or substance.

For example, in Judgment No.~5, two PoLs were detected and summarized. According to the notes, the first was paraphrased, while the second had some words replaced with synonyms. In these cases, although the PoLs were not copied verbatim, their meaning remained unchanged.

However, one instance revealed a critical error. Although the PoL was reported with 100\% word accuracy, its incipit was incorrectly modified. The original phrase, “The Court of Cassation established: ...”, was mistakenly reported as “The Constitutional Court established: ...”. While the quoted text itself was accurate, misattributing the PoL to a different court introduced a significant legal error.

This occurred in Judgment No.~26 and highlights that, in legal contexts, even a single incorrect word can fundamentally alter the meaning of a statement. While the core concept remained unchanged, misattributing the source carried significant legal implications, potentially more serious than a semantic modification. Consequently, this PoL was classified as a hallucination.

Although PoLs are generally reported accurately, a 99\% accuracy rate remains insufficient and cannot be considered fully reliable. In contexts where precision is critical, even minor modifications – such as the misattribution described above – cannot be dismissed as trivial errors. Such changes can significantly alter meaning and compromise the integrity and intent of the original text.

Additionally, even in judgments that are neither particularly long nor dense with PoLs, ChatGPT occasionally generates hallucinations. For example, in Judgment No.~24, as recorded in the analysis, it produced three PoLs, the median number extracted in previous cases. However, all three were entirely fabricated, with topics completely unrelated to the source text.  

Notably, the presence of hallucinated PoLs does not imply that all extracted PoLs are hallucinations. More concerning, the presence of correct PoLs does not guarantee the absence of hallucinated ones, as observed in Judgment No.~17.

PoLs are often repeated multiple times within judgments, emphasizing their significance. However, this repetition does not imply that a judgment is confined to a single legal topic. For example, some judgments may address both human rights and jurisdiction simultaneously.

Furthermore, running the same query twice – with the same prompt and judgment – may yield different results. This inconsistency underscores ChatGPT's lack of determinism in its responses. As a result, each judgment had to be processed individually, with a separate query for each. Uploading a single file containing all judgments would not have yielded any results, not even partially accurate ones.

We also observed that ChatGPT's display can vary over time. In some cases, it cites a source for the extracted text but incorrectly refers to a prior file rather than the one being analyzed. While the accuracy of the cited sources was not specifically verified, inconsistencies between the referenced document and the actual file under review were confirmed. In other cases, the attribution is correct, but the content itself is entirely fabricated.

The main concern is that false PoLs often appear highly convincing. They match the topic, follow the structure of a real PoL, and could easily be mistaken for an actual excerpt. The output may resemble a valid summary of the judgment or even a legitimate principle identified earlier. This makes hallucinations difficult to detect at a glance: spotting them requires a careful comparison with the original text. 

To mitigate this issue, we found it necessary to start a new chat every 5-7 queries. Otherwise, hallucinations and mix-ups with previously analyzed files became more frequent.

It is important to emphasize, as detailed in the dataset section, that all 60 analyzed judgments share similar technical features and thematic focus. The annotation process was conducted on Italian judgments, specifically rulings from first-instance and second-instance courts (Tribunals and Courts of Appeal), concerning LGBTQIA+ rights.

This similarity made it particularly challenging to immediately recognize hallucinations and distinguish them from mix-ups with previously analyzed rulings.

Notably, during the review of PoLs identified by ChatGPT, we found two implicit PoLs that ChatGPT detected but the annotators did not.

\begin{figure}[H] 
    \centering
    \includegraphics[width=1\linewidth]{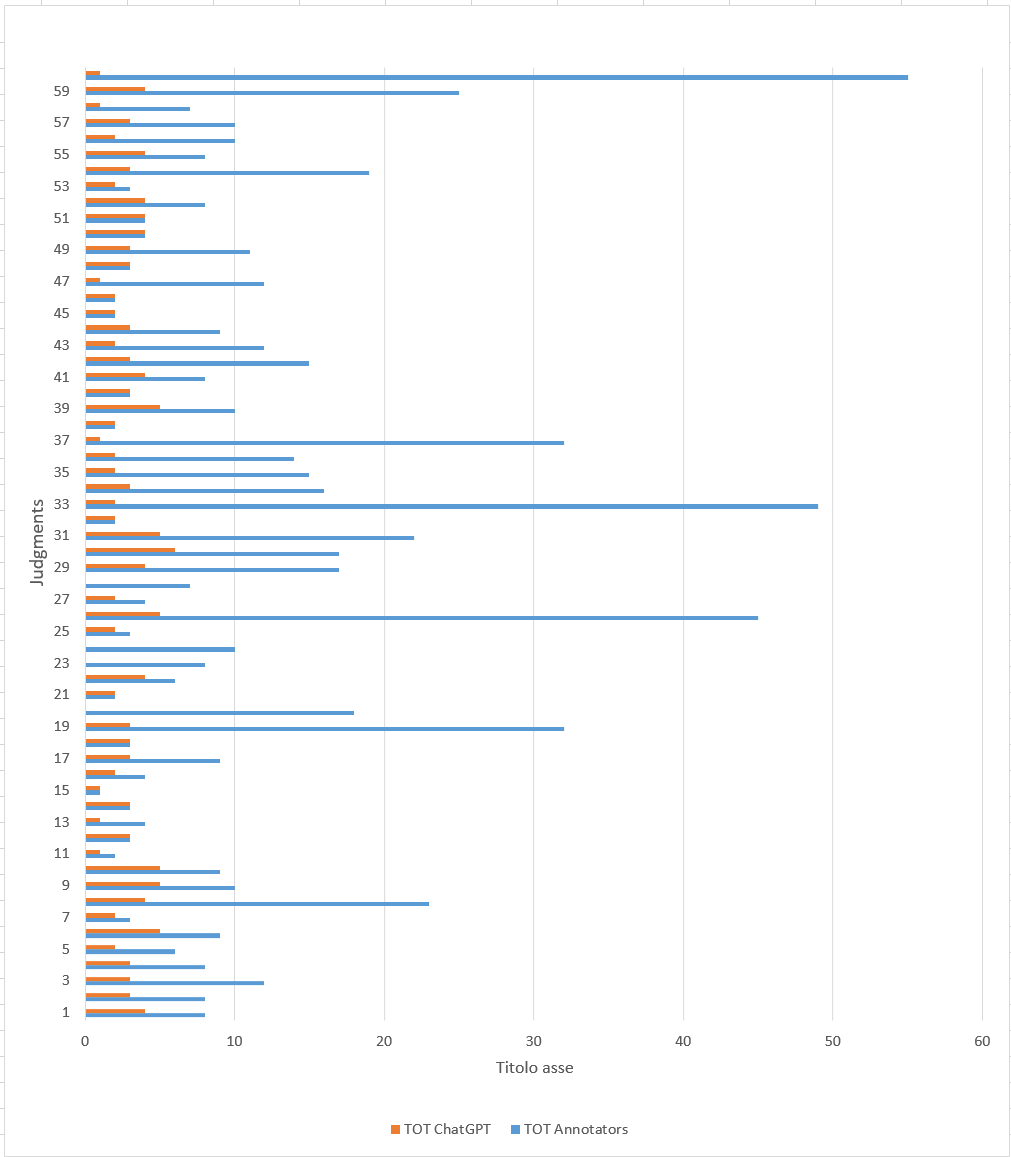}
    \caption{Comparison of the number of PoLs identified by the annotators versus those correctly detected by ChatGPT, with an average of 2.72, roundable up to 3. 
    }
    \label{G1_AvC}
\end{figure}

Figure \ref{G1_AvC} displays the number of PoLs identified in each decision, comparing those found by annotators (blue) with the correct ones detected by ChatGPT (orange). Hallucinations and incorrect PoLs generated by ChatGPT are here excluded.

As previously mentioned, this image also confirms that the query is most accurate when detecting a small number of PoLs and becomes less effective as the number increases.

\begin{figure*}
\centering
\includegraphics[width=1\textwidth]{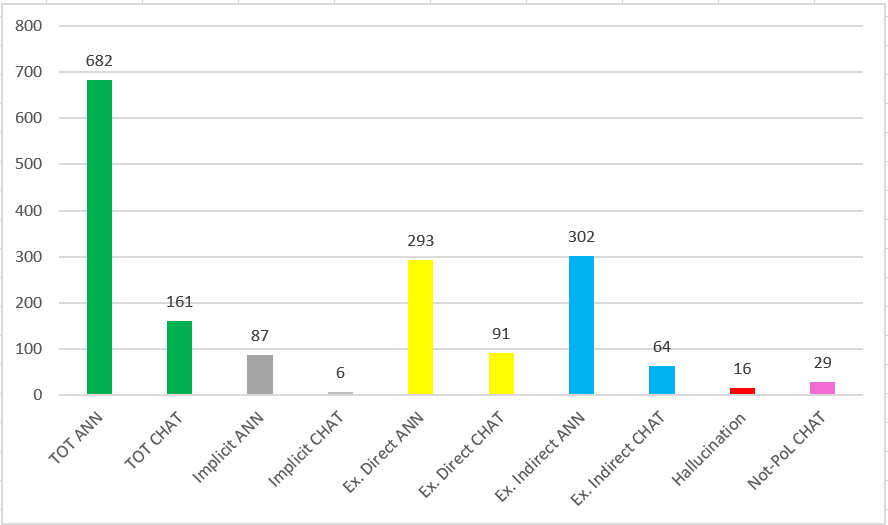}
\caption{Detailed comparison between the number of PoLs found by the annotators and by ChatGPT. Specifically, the comparison includes: the total number of correct PoLs found (labeled as ``\textit{TOT ANN}” for annotators and ``\textit{TOT CHAT}” for ChatGPT); implicit PoLs identified (labeled as ``\textit{Implicit ANN}” for annotators and ``\textit{Implicit CHAT}” for ChatGPT); explicit direct PoLs identified (labeled as ``\textit{Ex. Direct ANN}” for annotators and ``\textit{Ex. Direct CHAT}” for ChatGPT); explicit indirect PoLs identified (labeled as ``\textit{Ex. Indirect ANN}” for annotators and ``\textit{Ex. Indirect CHAT}” for ChatGPT). Then, hallucinations generated by ChatGPT and finally passages extracted as PoLs that were not actual PoLs (labeled as ``\textit{Not-PoL CHAT}").}
\label{G2_AvC}
\end{figure*}

The average number of PoLs detected is 2.72, which rounds to 3, highlighting a consistent pattern.

Figure \ref{G2_AvC} provides a detailed analysis of the experiment results, using the total values from each column in Table \ref{tabAnn} to highlight key comparisons. 

This figure compares the total number of correct PoLs identified by the annotators and ChatGPT, highlighting differences in the types of PoLs detected by each.

The gap between the total PoLs identified by the annotators and those detected by ChatGPT remains evident. As noted earlier, explicit PoLs (both direct and indirect) were easier to recognize due to their distinct semantic structure. Notably, ChatGPT also identified some implicit PoLs, likely by analyzing context, presentation style, or recurring keywords. However, since implicit PoLs are less standardized than explicit ones, they remain more challenging to detect.

The graph also highlights errors made by ChatGPT, including hallucinations and misidentified texts (“Not-PoL”). These errors often involve facts, statements by the parties (in quotes), or sections of the decision mistakenly classified as PoLs. While misclassifications are easy to spot, hallucinations require closer scrutiny, as they involve entirely fabricated information.

For a final interpretation of the data, it is important to emphasize that the wording of the decision is determined by the judge. As noted in the limitations section, the clarity of the original text plays a crucial role in the accurate extraction of PoLs.

\subsection{Results}

Table \ref{Chat:total} presents the results of PoLs extraction using ChatGPT for the 60 judgments.  
Therefore, at first, Table \ref{Chat:total} displays the total number of PoLs detected by ChatGPT and the mistakes made. In turn the mistakes are divided into two categories: Not-PoLs and Hallucinations.
\begin{table}[h]
\centering
\begin{tabular}{cccc}
    \textbf{Judgments}  & \textbf{PoLs}  & \multicolumn{2}{c}{\textbf{Errors}}\\
    \midrule
     60  & 161 & \multicolumn{2}{c}{45} \\
     & &  \textbf{Not-PoLs} &\textbf{Hallucinations}\\
     \cmidrule{3-4}
     & &  16 & 29 
 \end{tabular}
 \caption{PoLs Extracted by ChatGPT.}
 \label{Chat:total}
\end{table}

The types of PoLs are categorized as explicit, implicit, or indirect. The degree of completeness is classified into fully detected, partially detected, and partially detected with an ending of “(...)”, which may indicate a potentially full detection but is too long for ChatGPT's response. The degree of similarity is classified into three categories: PoLs that are identical to the original judgment, PoLs that are summarized while preserving the original meaning, and PoLs that are mostly the same but with some words changed. In the latter two cases, the meaning remains the same. However, if a word change alters the meaning, the PoL is classified as hallucinated.

The characteristics of the 161 PoLs are shown in Table~\ref{Chat:PoLs}. 

\begin{table}[htb!]
    \centering
    \begin{tabular}{ccc}
    \multicolumn{3}{c}{\textbf{Type of PoLs}}\\
    \textbf{Explicit} & \textbf{Implicit} & \textbf{Indirect}\\
    \midrule
    91 & 6 & 64\\[24pt]
    \multicolumn{3}{c}{\textbf{Degree of Completeness}}\\
    \textbf{Full} & \textbf{Partial} & \textbf{Partial “(...)”}\\
    \midrule
    91 & 61 & 9 \\[24pt]
    \multicolumn{3}{c}{\textbf{Degree of Similarity}}\\
    \textbf{Same Text} & \textbf{Summary} & \textbf{Word Exchange}\\
    \midrule
    154 & 4 & 5
    \end{tabular}
        \caption{Characteristric of the PoLs extracted by ChatGPT}
        \label{Chat:PoLs}
    \end{table}

Table \ref{tab:confusione} presents the confusion matrices, comparing the expert-led annotation results with those obtained from ChatGPT extraction.\footnote{The annotators identified 682 PoLs. During the experiment, ChatGPT detected 2 additional PoLs that the annotators had missed. In the third step, Regex detected 2 more PoLs that neither the annotators nor ChatGPT had identified.  Thus, the total number of PoLs is 686.}

\begin{table}[h]
    \centering
    \begin{tabular}{ccc}
    \toprule
     \multicolumn{3}{c}{Annotators}\\                 
                   &  Yes  & No\\
    \midrule
     PoLs          &  682  &   4\\
     not PoLs      &    0  &   0 \\
     \bottomrule
    \end{tabular}
\hspace*{20pt}
    \begin{tabular}{ccc}
    \toprule
     \multicolumn{3}{c}{ChatGPT}\\                       
                   &  Yes  & No\\
    \midrule
     PoLs          &   161 & 525 \\
     not PoLs      &    45 &   0 \\
     \bottomrule
    \end{tabular}
    \vspace*{12pt}
  \caption{Confusion Matrices Annotators - ChatGPT.}
   \label{tab:confusione}
\end{table}

The details are as follows: 

\begin{itemize}
    \item 161: Total number of PoLs correctly identified by ChatGPT.
    \item 2: Cases in which ChatGPT identified correct PoLs that were overlooked by the annotators.
    \item 525: PoLs missed by ChatGPT, calculated as the difference between the total PoLs and the number of PoLs correctly detected by ChatGPT.
    \item 45: Errors made by ChatGPT, including hallucinations and Not-PoLs it generated.
\end{itemize}
Based on the above figures, the precision, recall, accuracy, and F1-score for the experiment are as follows:
\begin{center}
\begin{tabular}{lcc}
    & Annotators & ChatGPT \\
    \midrule
\textbf{Precision} &  0.994  &  0.235 \\
\textbf{Recall}    &  1.0    &  0.781 \\
\textbf{Accuracy}  &  0.994  &  0.220 \\
\textbf{F1-score}  &  0.997  &  0.361
\end{tabular}
\end{center}

\section{Exploration with Regex}

The exploration with Regex, as well as the one with ChatGPT, was carried out in two main stages: (i) discovering regex and (ii) identifying principles.

\subsection{Discovering Regex}

As mentioned in the introduction, this study builds upon a preliminary investigation into the extraction of PoLs from court decisions using Regex. \cite{MariICAIL2023}.  

Given the limited quantity and quality of results obtained through extraction with ChatGPT, and considering that the Regex-based extraction developed by human IT experts in the previous study had yielded promising preliminary insights, we explored an alternative approach. Specifically, we sought to generate a regular expression for PoLs extraction directly using ChatGPT.

Building on the prompt experience gained from previous query submissions to ChatGPT for this purpose, this phase directly utilized the “Few-shot learning” approach. This method involved incorporating relevant examples into the prompt to improve the quality of the generated Regex \cite{yu2023exploring}.

More specifically, we first asked ChatGPT to generate a Python script using Regex to extract paragraphs containing text in quotation marks, a simple form of citation, and specific keywords.\footnote{The interaction with GPT was conducted with in VS Code, using the GitHub Copilot plugin configured to use GPT-4o. We also compared the Regex and Python scripts with a version generated by ChatGPT. While there were some syntactical differences, the Regex were the same and the Python scripts were essentially equivalent.}

The prompt we used is the following:
\begin{Verbatim}[fontsize=\small]
# write a python script that uses regular expressions and extracts 
# the paragraphs containing a sentence in quotation marks. 
# Quotations marks can be either english style or italian style or 
# french style.

# The script should be able to extract the paragraphs where there is
# a passage in quotation marks followed by a number in parenthesis.

# The script should be able to extract the paragraphs until a new 
# one where the words  CORTE, TRIBUNALE, GIURISPRUDENZA, COLLEGIO, 
# CONSESSO, CASSAZIONE even when abbreviated.
\end{Verbatim}
The Python script generated is presented below:
\begin{Verbatim}[fontsize=\small]
def extract_paragraphs(text):
    # Define the regular expression patterns
    quote_pattern = r'([“"«‘"].*?[”"»’"])'
    citation_pattern = r'\(.*?\d{4}\)'
    keyword_pattern = r'\b(CORTE|TRIBUNALE|TRIB\.|' \
    'GIURISPRUDENZA|COLLEGIO|CONSESSO|CASS\.|CASSAZIONE)\b'

    paragraphs = text.split('\n')
    extracted_paragraphs = []

    for paragraph in paragraphs:
        if re.search(quote_pattern, paragraph) or 
        re.search(citation_pattern, paragraph):
            extracted_paragraphs.append(paragraph)
        elif re.search(keyword_pattern,paragraph,re.IGNORECASE):
            extracted_paragraphs.append(paragraph)

    return extracted_paragraphs
\end{Verbatim}
The prompt was then refined, resulting in a second, improved prompt  combining the already genenerated Regex. 
Morover the Python script was expected to extract paragraphs containing PoLs from a pool of files (in .docx format), and generate a CSV file for each file containing the extracted paragraphs. The prompt was as follows:
\begin{Verbatim}[fontsize=\small]
# write a python function that takes as input a directory 
# called “Sentenze".  The directory contains a set of word files. 
# For each word file in the directory, parse the file paragraph by 
# paragraph, and extract the paragraphs matching either  
# 1) both the quote_pattern and the keyword_pattern; or
# 2) the citation_pattern when the citation_pattern is at the end 
#    of the paragraph

# for each file the function returns a  csv file with the same 
# basename as the input file. The csv contains 2 columns. The content 
# of the columns depends on what pattern was matched. For the case of 
# the quote_pattern and the keyword_pattern, the first column contains
# the paragraph and the second column contains the quote. For the case 
# of the citation_pattern, the first column contains the paragraph and 
# the second column is empty. 

# the function should save the csv files in a new directory called 
# "Principi"
\end{Verbatim}
The Python script generated is presented below:
\begin{Verbatim}[fontsize=\small]
def extract_paragraphs_from_directory(directory):
    quote_pattern = r'([“"«‘"].*?[”"»’"])'
    citation_pattern = r'\(.*?\d{4}\)$'
    keyword_pattern = r'\b(CORTE|TRIBUNALE|TRIB\.|' \
    'GIURISPRUDENZA|COLLEGIO|CONSESSO|CASS\.|CASSAZIONE)\b'

    output_directory = os.path.join(directory, 'Principi')
    os.makedirs(output_directory, exist_ok=True)

    for filename in os.listdir(directory):
        if filename.endswith('.docx'):
            doc_path = os.path.join(directory, filename)
            doc = docx.Document(doc_path)
            extracted_data = []

            for paragraph in doc.paragraphs:
                text = paragraph.text
                if re.search(quote_pattern, text) and 
                re.search(keyword_pattern, text, re.IGNORECASE):
                    quote = re.search(quote_pattern, text).group(0)
                    extracted_data.append([text, quote])
                elif re.search(citation_pattern, text):
                    extracted_data.append([text, ''])

            csv_filename = os.path.splitext(filename)[0] + '.csv'
            csv_path = os.path.join(output_directory, csv_filename)

            with open(csv_path, mode='w', newline='', 
                      encoding='utf-8') as csv_file:
                writer = csv.writer(csv_file)
                writer.writerow(['Paragraph', 'Quote'])
                writer.writerows(extracted_data)

extract_paragraphs_from_directory('Sentenze')   
\end{Verbatim}

Thus, the Python script was then applied using a text editor. The complete analysis of the PoL extraction is outlined in the next section.

\smallskip

Although ChatGPT successfully generated the above mentioned regular expressions and a script to extract PoLs, it was unable to directly apply them to the provided judgments, as it admitted. In fact, when asked to \textit{Find in the attached file the portions of text that match the Regex you created}, it answered as follows: 

\begin{itemize}
    \item \textit{I did not find any matches when applying the pattern to the example sentence.}
    \item \textit{(...) This suggests that the automatic extraction may not have worked correctly due to the PDF encoding or a protection that prevents direct text extraction.}
    \item \textit{Despite attempts to clean and optimize the text, there are still no useful results with the applied Regex. This suggests that the format of the document extracted from the PDF may not contain typical references, or the required legal citations might be represented in a way different from expectations. At this point, I suggest proceeding with a manual review of the file to accurately identify the format of the legal references.}
\end{itemize}

\subsection{Identifying PoLs}

Each of the 60 judgments in the dataset underwent this process. The extracted results were then compared with those manually annotated by human experts to assess the accuracy of the regex-based extraction.

Two experts conducted this evaluation, as in the previous phase.

The first expert performed the searches in the text editor, applying the ChatGPT-generated Regex to each .docx judgment, reviewing the extracted text, and reading the results aloud.

The second expert cross-checked the findings by referring to the original annotated documents. Their task was to verify whether the PoLs identified by the Regex, as read aloud by the first expert, accurately corresponded to those manually annotated. They also recorded the results in an .xlsx file, following the assessment criteria outlined in the next section.

\section{Analysis and Results: Annotators vs Regex}

\subsection{Analysis}

\begin{figure*}[ht!]
\centering
\includegraphics[width=1\textwidth]{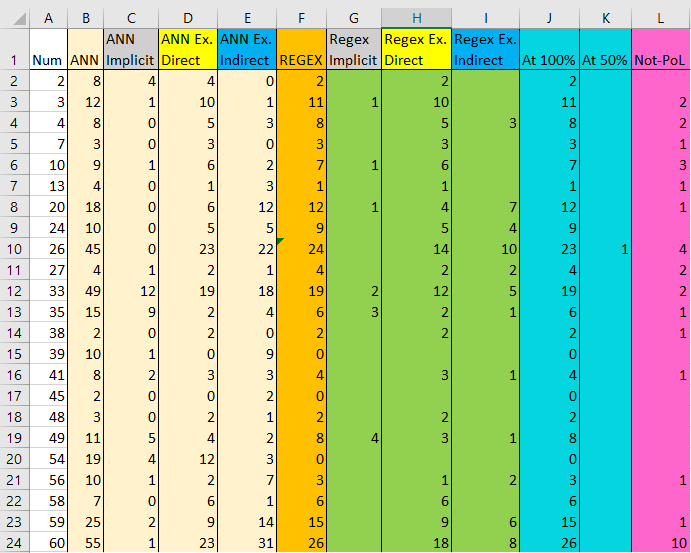}
\caption{Excerpt of the annotation table tracking Regex's PoLs extraction.}\label{tabreg}
\end{figure*}

Figure \ref{tabreg} provides an excerpt of the table used to track the annotation at object, showing  a comparison between the PoLs identified by the annotators and those detected by the Regex. The dataset remained consistent, with the same order applied to the expert annotators, ChatGPT, and Regex.

The columns B-E are identical to those in Figure \ref{tabAnn}, as they show the results from the expert annotators. The orange column (F) represents the total number of PoLs detected by the Regex, while the following three green columns (G-I) provide details on the types of PoLs: implicit, explicit-direct, and explicit-indirect. The two blue columns (J and K) indicate whether the PoL is found fully or partially. Finally, the last column lists concepts detected as PoLs that are not actually PoLs.

Overall, the Regex extraction performs better than ChatGPT's. All PoLs, except for one case (Judgment n.~26), were detected fully. However, in some instances, Regex did not detect any PoLs, such as in Judgments n.~39, 45, and 54.

As expected from the structure of a Regex, explicit-direct PoLs were more frequently identified. However, unexpectedly, implicit PoLs were also detected. Additionally, unlike ChatGPT, there is no average or fixed limit on the number of PoLs detected. The number of extractions per judgment ranges from 1 to 26, as seen in Judgments n.~13 and 60.

In general, there are few cases where the total number of PoLs detected by the annotators matches exactly with those detected by Regex (such as cases n.~4, 7, 27, and 38). However, the number of extractions by Regex is typically close to that of the annotators, whether the total PoLs are low (as in case n.~48), medium (as in cases n.~10 and 58), or high (as in cases n.~3, 24, and 49). Of course, there are cases where Regex detects only half (as in case n.~41) or even less than half of the PoLs (as in cases n.~2, 13, 35, and 56).

Curiously, in some cases, even though Regex detected fewer PoLs than the annotators, the number of PoLs it identified is still significant, especially considering the high number of PoLs detected by the annotators (as in cases n.~20, 26, 33, 59, and 60).

Moreover, the number of paragraphs identified as PoLs by the Regex-based methods has increased. However, many of these do not actually constitute PoLs (“Not-PoL”). In some cases, passages mistakenly identified as PoLs – whether enclosed in quotation marks or not – are actually references to legislative texts or excerpts from narratives.

\begin{figure*}[ht!] 
    \centering
    \includegraphics[width=1\linewidth]{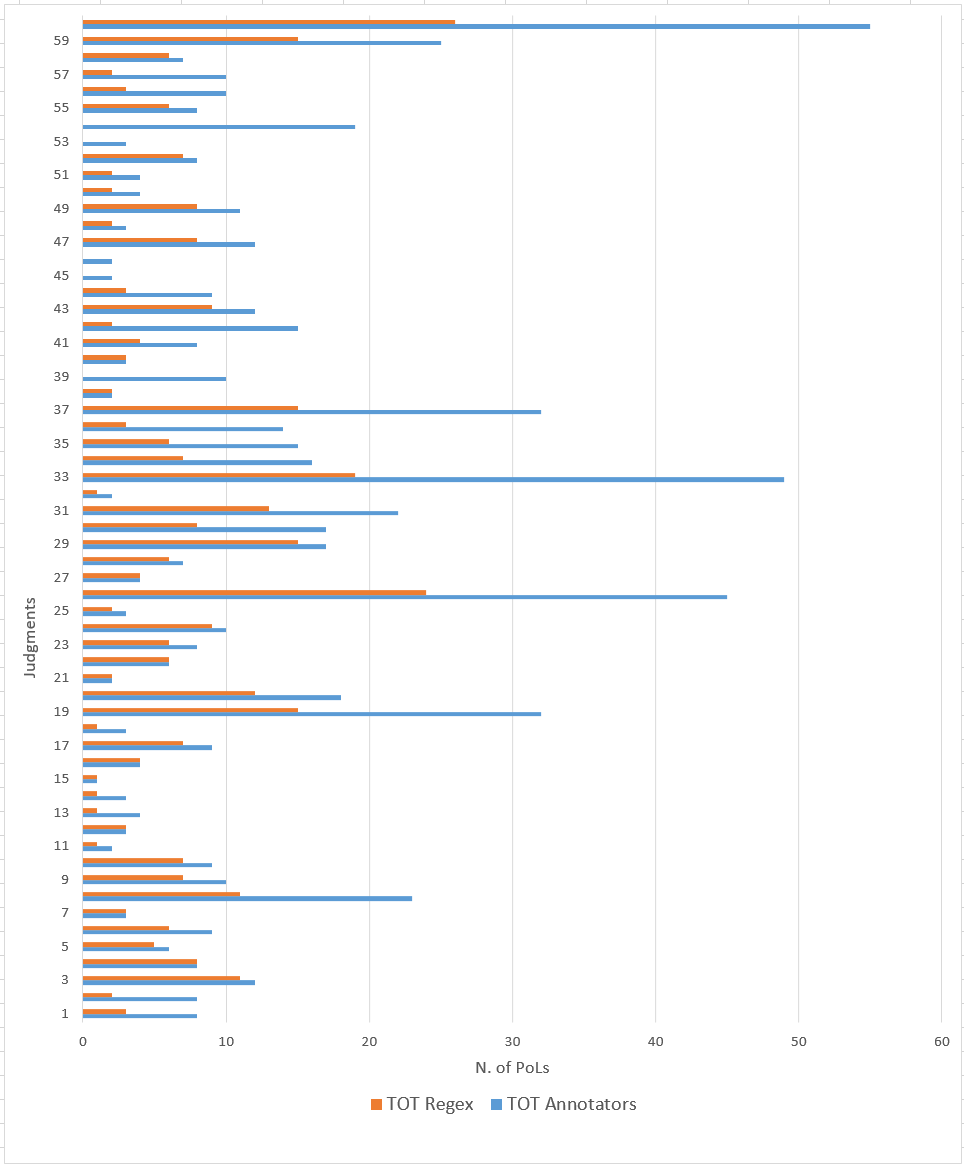}
    \caption{Comparison of the number of PoLs identified by the annotators versus those correctly identified by the Regex, with an average of 6.08. 
    }
    \label{G1_AvR}
\end{figure*}

Figure \ref{G1_AvR} presents a comparison between the PoLs identified by the annotators (in blue) and those correctly identified by Regex (in orange) for each of the 60 analyzed judgments. As mentioned earlier, the figure clearly shows that there are no recursion or limitations in PoLs detection, unlike with ChatGPT. Furthermore, the average number of PoLs detected is 6 per judgment, which is more than double the average detection rate of ChatGPT. 

\begin{figure*}
\centering
\includegraphics[width=1\textwidth]{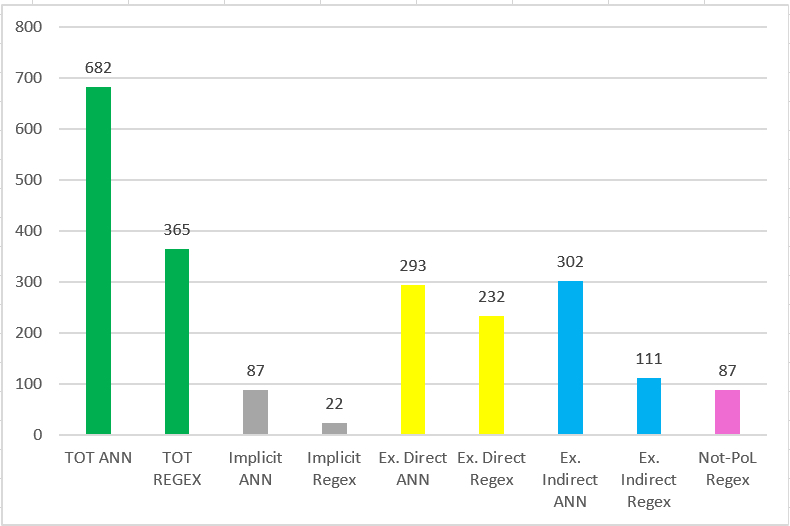}
\caption{Detailed comparison between the number of PoLs found by the annotators and by the Regex. Specifically, the comparison includes: the total number of correct PoLs found (labeled as ``\textit{TOT ANN}” for annotators and ``\textit{TOT REGEX}” for Regex); implicit PoLs identified (labeled as ``\textit{Implicit ANN}” for annotators and ``\textit{Implicit Regex}” for Regex); explicit direct PoLs identified (labeled as ``\textit{Ex. Direct ANN}” for annotators and ``\textit{Ex. Direct Regex}” for Regex); explicit indirect PoLs identified (labeled as ``\textit{Ex. Indirect ANN}” for annotators and ``\textit{Ex. Indirect Regex}” for Regex). Finally passages extracted as PoLs that were not actual PoLs (labeled as ``\textit{Not-PoL Regex}”).}
\label{G2_AvR}
\end{figure*}

Figure \ref{G2_AvR} provides a detailed analysis of the experiment results, starting with the total values from each column presented in Table \ref{tabAnn} for comparison. The graph displays the total number of correct PoLs identified by both the annotators and Regex. It also compares the different types of PoLs identified by each. Finally, it shows the number of errors made in Regex's detection (“Not-PoL”).

As with ChatGPT, explicit PoLs (both direct and indirect) were easier to detect due to their clear semantic characteristics. Interestingly, as anticipated, despite the structure of Regex, it was also able to identify a few implicit PoLs, as shown in the grey column.

Regarding errors, although Regex does not hallucinate, it extracted a considerable number of paragraphs that are not actually PoLs.
As anticpated, these are references to legislative texts or excerpts from narratives. Errors that are easily identifiable, in contrast to true hallucinations or entirely fabricated PoLs generated by ChatGPT.

\subsection{Results}
The results for the Regex extraction are presented in Table \ref{Regex:PoLs}.
\begin{table}[h]
    \centering
    \begin{tabular}{ccc}
       \textbf{Judgments}  & \textbf{PoLs}  &  \textbf{Not-PoLs}\\
       \midrule
        60  & 365 & 87        
    \end{tabular}
    \caption{PoLs Extracted by Regex.}
    \label{Regex:PoLs}    
\end{table}

As with the previous Table \ref{Chat:PoLs}, Table \ref{Regex:PoLs} presents the total number of PoLs found by Regex (365) and the number of PoLs identified as such but not actually PoLs (87), which are typically citation passages.

\begin{table}[h!]
    \centering
    \begin{tabular}{ccc}
    \multicolumn{3}{c}{\textbf{Type of PoLs}}\\
    \textbf{Explicit} & \textbf{Implicit} & \textbf{Indirect}\\
    \midrule
    232 & 22 & 111
    \end{tabular}
    \hspace*{40pt}
    \begin{tabular}{cc}
    \multicolumn{2}{c}{\textbf{Degree of Completeness}}\\
    \textbf{Full} & \textbf{Partial} \\
    \midrule
     364 & 1
    \end{tabular}
        \caption{Breakdown of PoLs extracted by Regex.}
        \label{Regex:type}
\end{table}

Table \ref{Regex:type} gives a breakdown of the 365 PoLs identified by Regex,
consisting of 22 implicit, 232 explicit-direct, and 111 explicit-indirect PoLs. It also shows the number of PoLs identified fully (364) and the one identified partially. 

The details of the extraction using Regex are as follows: 
\begin{itemize}
    \item 365: Total number of PoLs correctly identified by Regex.
    \item 2: Cases where Regex identified correct PoLs that were missed by the annotators.
    \item 321: PoLs missed by Regex, calculated as the difference between the total PoLs and those correctly identified by Regex.
    \item 87: Errors done by Regex, i.e. not-PoLs that it identified as PoLs.
\end{itemize}
Therefore, we present the following confusion matrix (and again include the confusion matrix for the Expert Annotators for comparison):
\begin{table}[h]
    \centering
    \begin{tabular}{ccc}
    \toprule
     \multicolumn{3}{c}{Annotators}\\                 
                   &  Yes  & No\\
    \midrule
     PoLs          &  682  &   4\\
     not PoLs      &    0  &   0 \\
     \bottomrule
    \end{tabular}
\hspace*{20pt}
    \begin{tabular}{ccc}
    \toprule
     \multicolumn{3}{c}{Regex}\\                       
                   &  Yes  & No\\
    \midrule
     PoLs          &   365 & 321 \\
     not PoLs      &    87 &   0 \\
     \bottomrule
    \end{tabular}
    \vspace*{12pt}
  \caption{Confusion Matrices Annotators - Regex. }
    \label{tab:confusioneReg}
\end{table}

Based on the above figures, the precision, recall, accuracy, and F1-score for the experiment are:
\begin{center}
\begin{tabular}{lcc}
    & Annotators & Regex \\
    \midrule
\textbf{Precision} &  0.994  &  0.532 \\
\textbf{Recall}    &  1.0    &  0.807 \\
\textbf{Accuracy}  &  0.994  &  0.472 \\
\textbf{F1-score}  &  0.997  &  0.641
\end{tabular}
\end{center}

\section{Comparison: Annotators vs ChatGPT vs Regex}

At the conclusion of this work, we aim to present a comparison of the three extractions (made by the expert annotators, ChatGPT, and Regex) to highlight key observations.

The comparison and the related results suggest that, while ChatGPT successfully generated the required regular expressions, it struggled to apply them effectively, as can be seen in the graphical details provided below.

\begin{figure*}[ht!]
\centering
\includegraphics[width=1\textwidth]{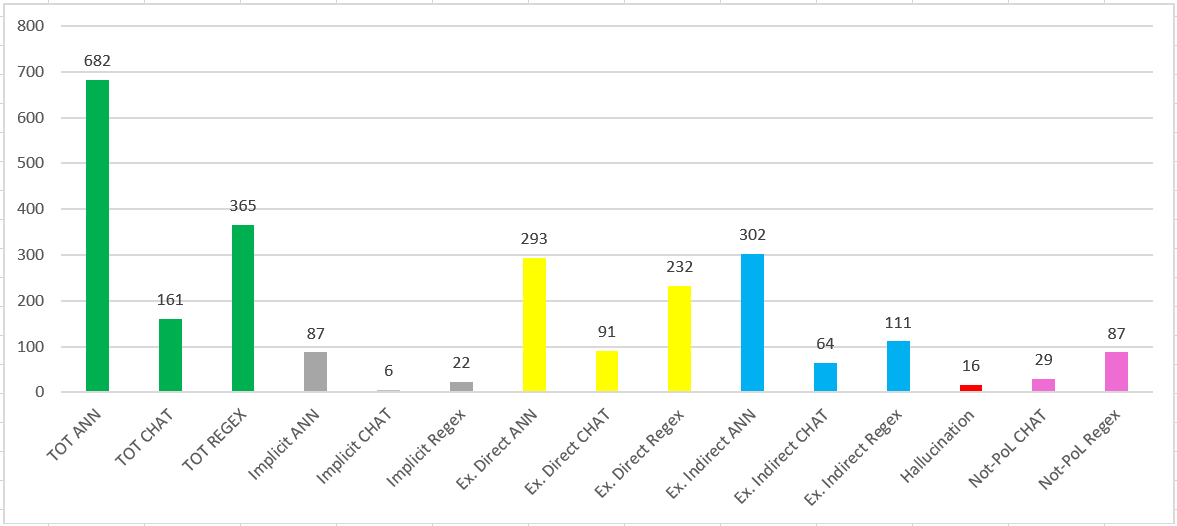}
\caption{Detailed comparison between the number of PoLs found by the annotators, ChatGPT and the Regex. Specifically, the comparison includes: the total number of correct PoLs found (labeled as ``\textit{TOT ANN}” for annotators, ``\textit{TOT CHAT}” for ChatGPT and ``\textit{TOT REGEX}” for Regex); implicit PoLs identified (labeled as ``\textit{Implicit ANN}” for annotators, ``\textit{Implicit CHAT}” for ChatGPT and ``\textit{Implicit Regex}” for Regex); explicit direct PoLs identified (labeled as ``\textit{Ex. Direct ANN}” for annotators, ``\textit{Ex. Direct CHAT}” for ChatGPT and ``\textit{Ex. Direct Regex}” for Regex); explicit indirect PoLs identified (labeled as ``\textit{Ex. Indirect ANN}” for annotators, ``\textit{Ex. Indirect CHAT}” for ChatGPT and ``\textit{Ex. Indirect Regex}” for Regex). Finally, error passages extracted as hallucinations (labeled as ``\textit{Hallucination}” for ChatGPT) and PoLs that were not actual PoLs (labeled as  ``\textit{Not-PoL CHAT}” for ChatGPT and ``\textit{Not-PoL Regex}” for Regex).}
\label{G2_AvCvR}
\end{figure*}

Figure \ref{G2_AvCvR} presents a graphic comparison of the results from the expert annotators, ChatGPT, and Regex, showing the total number of PoLs and providing details about the types of PoLs (implicit, explicit-direct, and explicit-indirect), as well as the errors (hallucinations and not-PoLs).

\begin{table}[hb!]
    \centering
    \begin{tabular}{l|l|lll}
             & \textbf{PoLs} & \textbf{Implicit} & \textbf{Explicit} & \textbf{Indirect} \\ 
             \midrule
\textbf{Whole PoLs}& 686& 91&293&302\\ 
\textbf{Annotators} & 682 (99.4\%)&87 (95.6\%)&293 (100\%)&302 (100\%)\\ 
\textbf{ChatGPT}& 161 (23.5\%)&6 (6.6\%)&91 (31\%)&64 (21.2\%) \\ 
\textbf{Regex}& 365 (53.2\%)& 22 (24.2\%)&232 (79.2\%)&111 (36.7\%) 
    \end{tabular}
    \caption{Comparison between PoLs founded by Annotaters, ChatGPT and Regex.} 
    \label{VS}
\end{table}

Table \ref{VS} presents the results of the comparison, showing the numbers and percentages of PoLs identified by the expert annotators, ChatGPT, and Regex. It summarizes the results from Tables \ref{tabAnn}, \ref{Chat:total}, and \ref{tabreg}. Like the previous tables, it shows the total number of PoLs identified, along with details on how many are implicit, explicit-direct, and explicit-indirect. The percentages are calculated based on the “Whole PoLs” row to illustrate the weight of each type of PoL. As expected, the human annotators' work is by far the most accurate, and the results differ significantly from both tools. However, since annotators are not perfect, both tools contributed to improving the work by identifying 2 PoLs each.

Regex performs significantly better than ChatGPT, both in terms of total PoLs and across the different types of PoLs. In general, Regex identifies 1 PoL out of every 2, while ChatGPT identifies only 1 PoL out of every 5.

It can also be observed that explicit-direct PoLs are the easiest to identify for all methods, due to their clear structure. They are followed by explicit-indirect PoLs, which, while still structured, are slightly more challenging. Finally, implicit PoLs are the most difficult to detect.

It should be noted that implicit PoLs are the most difficult to identify, not only for the tools but also for the annotators. This is due to their structure, where the principle is paraphrased and no citations are provided.

This occurs because implicit principles are generally more established than others - they represent concepts that have become deeply embedded in society and in the prevailing notion of justice. Over time, these principles are cited less frequently, as referencing them would become redundant and overwhelming due to the sheer number of instances in which they apply. As a result, identifying them becomes particularly challenging for a machine, which lacks both a sense of justice and an inherent moral compass. In contrast, when a human fails to recognize such a principle, it may simply indicate that the notion is so deeply ingrained in their understanding that they no longer perceive it as a distinct principle but rather as a fundamental concept.

In other words, relying solely on the manually extracted principles or depending entirely on the tool would result in these inherent concepts being overlooked. They would be excluded rather than integrated into a broader notion of common sense, whether for the reader or the AI.

\begin{figure*}
\centering
\includegraphics[width=1\textwidth]{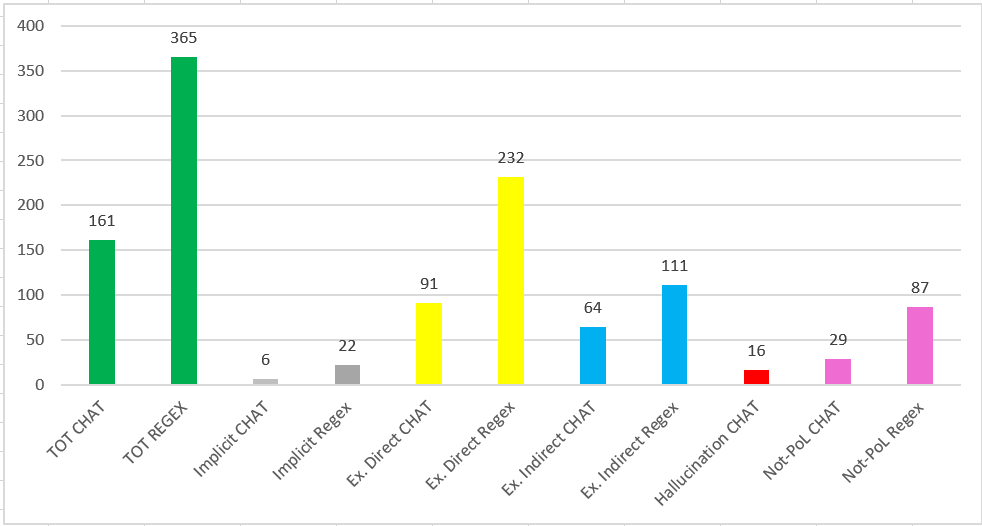}
\caption{Detailed comparison between the number of PoLs found by ChatGPT and the Regex. Specifically, the comparison includes: the total number of correct PoLs found (labeled as ``\textit{TOT CHAT}” for ChatGPT and ``\textit{TOT REGEX}” for Regex); implicit PoLs identified (labeled as ``\textit{Implicit CHAT}” for ChatGPT and ``\textit{Implicit Regex}” for Regex); explicit direct PoLs identified (labeled as ``\textit{Ex. Direct CHAT}” for ChatGPT and ``\textit{Ex. Direct Regex}” for Regex); explicit indirect PoLs identified (labeled as ``\textit{Ex. Indirect CHAT}” for ChatGPT and ``\textit{Ex. Indirect Regex}” for Regex). Finally, error passages extracted as hallucinations (labeled as ``\textit{Hallucination CHAT}” for ChatGPT) and PoLs that were not actual PoLs (labeled as ``\textit{Not-PoL CHAT}” for ChatGPT and ``\textit{Not-PoL Regex}” for Regex). }
\label{G2_CvR}
\end{figure*}

To emphasize the difference between the results of ChatGPT and Regex, Figure \ref{G2_CvR} focuses solely on the graphic comparison between the two in terms of total PoLs detected, type of PoLs and errors made.

\begin{table}[H]
    \centering
    \begin{tabular}{lcccc}
             & \textbf{PoLs} & \textbf{Errors} & \textbf{Not-PoLs}& \textbf{Hallucinations}\\ 
             \midrule           
       \textbf{ChatGPT}& 206& 45 (21.8\%)& 29 (14\%)& 16 (7.8\%) \\ 
       \textbf{Regex}  & 452& 87 (19 \%)& 87 (19 \%)&
    \end{tabular}
    \caption{Comparison between wrong PoLs found by ChatGPT and Regex.}
    \label{errors}
\end{table}

Table \ref{errors} zooms in further on the mistakes made by ChatGPT and Regex, specifically highlighting the sum of the errors and the detailed Not-PoLs and hallucinations. It shows the total PoLs identified by each tool, regardless of whether they were correct or incorrect. The related percentages are calculated based on these values\footnote{The percentage weight is calculated on the total PoLs identified by each tool. Thus, Tot\_PoLs\_ChatGPT = Correct\_PoLs\_ChatGPT + Wrong\_PoLs\_ChatGPT, and Tot\_PoLs\_Regex = Correct\_PoLs\_Regex + Wrong\_PoLs\_Regex.}.

Due to the weight of the percentage, it is possible to observe that, although the number of Not-PoLs identified by Regex is numerically high, its proportion relative to the PoLs found is lower compared to the errors made by ChatGPT.

Another important point is that although Not-PoLs are not actual PoLs (and thus have been counted as errors), they are parts of text, such as quotations from laws or narratives provided by the parties involved. These are easily recognizable by a human, especially by a legal expert. We can consider them as unnecessary but easily identifiable and discarded.

The major issue lies in hallucinations, as they are structured like true PoLs (either explicit-direct or indirect). However, they are not easily identifiable at a glance, even by an expert, unless the entire sentence is read, at which point the tool's usefulness would be diminished.

Based on the above figures and the confusion matrices in Table \ref{tab:confusione} and Table \ref{tab:confusioneReg}, the precision, recall, accuracy, and F1-score for the experiment are:

\begin{center}
\begin{tabular}{lccc}
    & Annotators & ChatGPT &  Regex \\
    \midrule
\textbf{Precision} &  0.994  &  0.235 & 0.532\\
\textbf{Recall}    &  1.0    &  0.781 & 0.807\\
\textbf{Accuracy}  &  0.994  &  0.220 & 0.472\\
\textbf{F1-score}  &  0.997  &  0.361 & 0.641
\end{tabular}
\end{center}

\section{Is AI Truly Intelligent?}

A common sign of intelligence is the ability to plan, break down complex tasks into simpler ones, and integrate them effectively \cite{correa2023humans}.

The final results clearly show that a simple Regex outperforms ChatGPT. Moreover, since ChatGPT generated the Regex, which was then used in a Python script to extract PoLs, this suggests that ChatGPT's intelligence is limited.

ChatGPT possesses the necessary information to perform better but lacks the ability to assemble it, reason through it, and integrate its knowledge effectively for a superior result. While the knowledge exists, human intelligence is required to guide the process and connect the steps coherently.

Artificial intelligence lacks true understanding and reasoning, making it fundamentally limited. Real intelligence remains a uniquely human trait.

The bigger issue is hallucinations, which are especially problematic because they are not easily identifiable at a glance - even for experts - without reading the entire legal document. This defeats the purpose of using AI as a support tool, such as for extracting PoLs from a judgment.

\section{Conclusions and Future Works}

This paper evaluates ChatGPT's ability to extract PoLs, comparing its performance against regular expressions as a baseline, and highlighting its limitations.

The baseline approach involved using ChatGPT to generate a Regex for extracting PoLs from court rulings, which was then applied using a text editor. The results from these two consecutive steps outperformed ChatGPT's independent performance in extracting PoLs from court rulings. This highlights that, at least for now, AI is not as advanced as it may seem.

Although AI performs well in general domains, it does not achieve the same level of success in specialized fields like law. This shows that, as of now, legal expertise cannot be replaced by AI tools, even for relatively straightforward tasks such as identifying PoLs. Expectedly, while extraction using Regex produced better results than extraction with ChatGPT, it still did not deliver exceptional outcomes. This highlights that, in this particular case, the performance of domain experts cannot be matched by any form of automation.

Thus, it is important to note that achieving a truly intelligent AI capable of operating in the legal domain would require further development and implementation.

Therefore, we propose the methodology behind the presented baseline as a potential pipeline that legal experts - without specific IT skills - can use to accelerate the identification of PoLs through automated extraction. This approach is user-friendly, requiring no specialized IT knowledge, and can be easily accessed by any average computer user. To facilitate this, the Regex provided in section 6.1 can be directly applied to a relevant dataset by following the steps outlined in section 6.2. These automated steps are simple, easily replicable, and can be repeated as part of a straightforward procedure accessible to any legal expert. Furthermore, the text editor used - like many other available tools - is freely accessible online, with no need for a subscription or local installation.

Future research will build on this analysis by assessing a dataset containing judgments on various legal topics (e.g., separation and divorce, inheritance, filiation) to gain a clearer understanding of any inaccuracies. We also plan to evaluate alternative large language models (LLMs), such as Deepseek, Lama, and OpenAI API, to explore potential improvements in performance.

Using different LLMs offers several advantages, including the ability to adjust query temperature (e.g., setting it to 0 via the OpenAI API), helping to reduce hallucinations. Unlike ChatGPT, which uses customized prompts and built-in safeguards that can influence responses, these models offer a more neutral and controlled testing environment.

Additionally, we plan to experiment with structured or multi-step prompting techniques. For example, breaking judgments into smaller units, such as paragraphs, may enhance accuracy, especially when dealing with lengthy legal texts. In our trial, ChatGPT processed a 38-page judgment containing 55 PoLs but successfully identified only 10, some of which were incorrect. Adopting a more structured approach could help overcome these limitations.

Based on the results, we also intend to conduct a comparison using NLP tools.

\bibliographystyle{elsarticle-num} 
\bibliography{biblio}

\appendix
\section{ChatGPT Prompting Strategy - Italian Version}
`Estrai i paragrafi in cui c’è la parola CORTE, TRIBUNALE, GIURISPRUDENZA, COLLEGIO, CONSESSO, CASSAZIONE e simili seguita da un passaggio tra virgolette. Esempio: La stessa Corte di Cassazione, pronunciandosi a Sezioni Unite, ha di recente affermato che si tratta di casi ``che interrogano profondamente la coscienza individuale e collettiva, ponendo questioni delicate e complesse, suscettibili di soluzioni differenziate”. Anche i paragrafi in cui c’è un passaggio tra virgolette seguito da un numero tra parentesi. Esempio: Si tratta di casi ``che interrogano profondamente la coscienza individuale e collettiva, ponendo questioni delicate e complesse, suscettibili di soluzioni differenziate” (cfr.~Cass.~S.U. Civili n.~12193/19). Anche i paragrafi fino al punto a capo in cui figurano le parole ‘la giurisprudenza ha sostenuto che…’, ‘come la corte ha statuito…’, ‘affermata giurisprudenza…’, ‘il principio stabilito dalla corte…’, seguiti o preceduti da un numero. Esempi: - La Corte Costituzionale, nella sentenza n.~120/2001, ha chiaramente affermato che il nome inteso come il primo ed immediato segno distintivo, costituisce uno dei diritti inviolabili della persona protetti dalla Carta ex art.~2 Cost., cui si riconosce il carattere di clausola aperta, con conseguente possibilità di evincere, dalla lettura combinata dell'art.~6 c.c., comma 3, e degli artt.~2 e 22 Cost., la natura di diritto soggettivo insopprimibile della persona. - Il riconoscimento del primario diritto alla identità sessuale, sotteso alla disposta rettificazione dell'attribuzione di sesso, rende consequenziale la rettificazione del prenome, che non va necessariamente convertito nel genere scaturente dalla rettificazione, dovendo il giudice tener conto del nuovo prenome, indicato dalla persona, pur se del tutto diverso dal prenome precedente, ove tale indicazione sia legittima e conforme al nuovo stato (Cass. Civ.~3877/2020). - Le spese della ctu, nella misura liquidata con separato decreto e operata la dimidiazione prevista dall'art.~130 tusg, vanno poste a carico dell'Erario (Corte Cost. 217/2019). - La Corte, dopo aver affermato che il legislatore […], aggiunge: “il legislatore […]”. Anche tutte le frasi che prima del punto terminano con un numero tra parentesi. Esempi: ‘…(Cass.~Civ. 3877/2020).’ Oppure: ‘…(Corte Cost.~217/2019).’ COPIA PEDISSEQUAMENTE I PASSAGGI DAL SINGOLO FILE CARICATO. ESPORTA PEDISSEQUAMENTE COSI’ COME SONO DAL SINGOLO FILE CARICATO. NON INVENTARE. NON RIASSUMERE. NON ASSEMBLARE. Segui dettagliatamente le istruzioni'.

\end{document}